\documentclass[runningheads]{llncs}

\usepackage[review,year=2024,ID=11513]{eccv}

\usepackage{eccvabbrv}
\usepackage{colortbl}
\usepackage{graphicx}
\usepackage{booktabs}
\usepackage{color}
\usepackage{threeparttable}
\usepackage{multirow}
\usepackage{ulem}
\usepackage[accsupp]{axessibility}  

\usepackage[pagebackref,breaklinks,colorlinks,citecolor=eccvblue]{hyperref}

\usepackage{orcidlink}

\begin{document}

\title{DDN-SLAM: Real-time Dense Dynamic Neural Implicit SLAM} 

\titlerunning{Abbreviated paper title}

\author{Li Mingrui\inst{1}\and
Yiming Zhou \inst{2} \and
Guangan Jiang\inst{1}\and
Tianchen Deng\inst{3}\and
Yangyang Wang\inst{4}\and
Hongyu Wang \inst{1}}

\authorrunning{F.~Author et al.}

\institute{Dalian University of Technology \and
Saarland University of Applied Sciences \and 
Shanghai Jiao Tong University \and 
Dalian Maritime University  \\
}

\maketitle

\begin{abstract}
SLAM systems based on NeRF have demonstrated superior performance in rendering quality and scene reconstruction for static environments compared to traditional dense SLAM. However, they encounter tracking drift and mapping errors in real-world scenarios with dynamic interferences. To address these issues, we introduce DDN-SLAM, the first real-time dense dynamic neural implicit SLAM system integrating semantic features. To address dynamic tracking interferences, we propose a feature point segmentation method that combines semantic features with a mixed Gaussian distribution model. To avoid incorrect background removal, we propose a mapping strategy based on sparse point cloud sampling and background restoration. We propose a dynamic semantic loss to eliminate dynamic occlusions. Experimental results demonstrate that DDN-SLAM is capable of robustly tracking and producing high-quality reconstructions in dynamic environments, while appropriately preserving potential dynamic objects. Compared to existing neural implicit SLAM systems, the tracking results on dynamic datasets indicate an average 90\% improvement in Average Trajectory Error (ATE) accuracy.
\end{abstract}

\section{Introduction}
\label{sec:intro}
In the fields of robotics and virtual reality, the Neural Radiance Fields (NeRF) \cite{guo2022nerfren, yu2021pixelnerf, mildenhall2021nerf, chen2021mvsnerf, xu2022point, barron2021mip, deng2023prosgnerf} have shown tremendous potential. There is a significant focus on real-time dense SLAM (Simultaneous Localization and Mapping) systems \cite{newcombe2011dtam, schops2019bad, bloesch2018codeslam, czarnowski2020deepfactors, newcombe2011kinectfusion, huang2021di} to combine with NeRF. Dense reconstruction \cite{chen2021mvsnerf, deng2022depthsupervisednerf, kerbl2023gaussian} is crucial for scene perception and understanding, providing improvements over sparse reconstruction \cite{qin2018vinsmono, wang2017stereodso, ling2017building, Qin2018, Hosseinzadeh2019, xie2022robust}. While traditional SLAM systems have achieved remarkable results in tracking and sparse reconstruction \cite{mur2017orb, yu2018ds}, low-resolution and discontinuous surface reconstruction fails to meet the requirements \cite{Whelan2015, McCormac2017, li2022bnv}. Recently, neural implicit SLAM methods \cite{zhu2022nice, wang2023coslam, yang2019cubeslam, chung2023orbeezslam, Li2023, deng2023plgslam} have demonstrated excellent performance, with systems capable of real-time reconstruction \cite{sucar2021imap, rosinol2022nerf, ming2022idf} and inference in large-scale virtual/real scenes. These SLAM systems outperform traditional SLAM methods in terms of texture details, memory consumption, noise handling, and outlier processing.
\noindent \par Although current neural implicit SLAM systems have achieved good reconstruction results in static scenes \cite{rosinol2022nerf, kong2023vmap, yang2022voxfusion, zhu2023nicerslam}, many real-world environments are often affected by dynamic objects, especially in applications such as robotics or autonomous driving, which involve complex physical environments and may also have low-texture areas or significant changes in lighting and viewing angles. Current neural implicit SLAM systems are unable to achieve effective tracking and reliable reconstruction in such environments. The depth information and pixel interference caused by dynamic objects lead to inaccurate tracking results, while severe occlusions result in distorted reconstructions and ghosting artifacts. In addition, traditional dynamic semantic SLAM systems
\cite{zhang2020flowfusion, whelan2015elasticfusion, he2023ovdslam, lv2018learning, whelan2012kintinuous} often treat potential dynamic objects as dynamic objects to be removed, resulting in holes. In the real world, foreground objects segmented by semantics often lie between dynamic and static states, rather than constantly in motion. These objects in potential motion do not interfere with the operation of the SLAM system and are of significant importance for the complete reconstruction of the scene. Traditional semantic SLAM systems also struggle to effectively differentiate between high-dynamic objects (involving significant human movement and high speed) and low-dynamic objects (involving minimal human movement and slower speed) and adopt different mapping strategies for them. 
\noindent \par To address these challenges, we propose DDN-SLAM, which is able to accurately differentiate between dynamic, static, and potentially static objects, instead of simply removing all foregrounds segmented by semantics. It utilizes the prior information provided by the semantic framework to constrain the segmentation range of dynamic object masks and performs feature point segmentation based on a mixture of Gaussian distributions. To improve the robustness of segmentation, we propose a reprojection error check to provide long-term data association, restoring features points that have been incorrectly removed. To accurately preserve potential static objects and achieve a complete reconstruction of the scene, we propose a mixed background restoration and rendering strategy. To fully utilize the results of static segmentation, we use map points guided by static sparse point cloud mapping to guide ray sampling. To prevent excessive removal, we use optical flow to distinguish low dynamic foreground constructed from sparse static feature points for background restoration. To constrain the impact of high-dynamic occlusion on rendering, we propese the motion consistency, depth, and color losses. Compared to other neural implicit SLAM systems\cite{zhu2022nice, yang2019cubeslam, chung2023orbeezslam}, we achieve a balance between real-time performance, low memory consumption, and high-quality geometric and texture details. In contrast to traditional dense semantic SLAM methods, our approach incorporates potential static objects into the scene representation, leading to high-quality rendering results. Our method can stably track and reconstruct in dynamic and challenging scenarios at a speed of 20Hz, supporting monocular, stereo, and RGB-D inputs.
In summary, our contributions can be summarized as follows:
\begin{itemize}
    \item We propose DDN-SLAM, the first dynamic SLAM system based on semantic features. By combining semantic detection framework and a Gaussian Mixture Model background depth probability check method, we distinguish foreground from background to eliminate interfering dynamic feature points. Long-term data association is established through a secondary check of reprojection errors of dynamic points and adjusting the Bundle Adjustment (BA) process. 
    
    \item We propose a mixed background restoration and rendering strategy, which includes using feature-based optical flow to differentiate dynamic masks for background restoration. We also propose a sampling strategy guided by static sparse point clouds to enhance the reconstruction of the static surface.
    
    \item We propose a rendering loss based on dynamic masks, including motion consistency loss, depth loss, and color loss, to constrain rendering artifacts of dynamic objects and remove occlusions.
\end{itemize}

\section{Related works}

\subsection{Traditional Dynamic Visual SLAM}
In most SLAM systems \cite{pire2017s, schops2019bad, newcombe2011dtam, mur2015orb}, eliminating false data from dynamic objects \cite{liao2022so, tian2022accurate} that are commonly present in the real world is an important problem. It holds significant implications for both camera tracking and dense reconstruction \cite{runz2017co, bloesch2018codeslam, craig2004tandem}.
Early researchers often used reliable constraints or multi-view geometric methods for feature matching, but they often failed to accurately segment subtly moving objects. Additionally, there was a lack of viable solutions for handling potential moving objects. Methods relying on optical flow consistency \cite{du2020accurate, soares2021crowdslam, wang2022drgslam, soares2019visual} could differentiate by detecting inconsistent optical flow generated by moving objects, but with lower accuracy.

In recent years, deep object detection networks that incorporate semantic information have been employed for pixel-level segmentation to more accurately deal with dynamic objects in real-world environments \cite{strecke2019fusion}. However, the depth models used in traditional SLAM methods often incur high computational costs, impacting real-time performance. Moreover, traditional methods suffer from an inability to perform scene representation and reasonable hole filling for complex textures and geometric details due to discrete or incomplete surface structures. Moreover, traditional dynamic semantic SLAM methods cannot accurately distinguish between dynamic, static, and potentially dynamic objects, leading to excessive removal after semantic segmentation, failing to preserve potentially dynamic objects that do not affect the mapping process.

\subsection{NERF-BASED SLAM}
The achievements of NeRF \cite{muller2022instant, wu2022d, wang2021neus, weng2022humannerf, turki2022mega, rebain2022lolnerf, mildenhall2021nerf} in 3D scene representation \cite{pearl2022nan, zakharov2020autolabeling, wang2022nerf, ichnowski2021dex, yu2022monosdf, hu2022nerf} have attracted widespread attention. In recent work, neural implicit SLAM systems and similar work \cite{lin2021barf, yao2019recurrent} have shown impressive performance in high-quality rendering and accurate 3D reconstruction. iMAP \cite{sucar2021imap} proposes a real-time updating system using a single MLP decoder, and NICE-SLAM \cite{zhu2022nice} extends the system to achieve larger scene representation, addressing the scene limitations of iMAP \cite{sucar2021imap}. ESLAM \cite{yang2019cubeslam} uses multi-scale vertically aligned feature planes for representation and decodes compact features directly into TSDF. Co-SLAM \cite{wang2023coslam} adopts a joint encoding method based on coordinates and one-blob to balance geometric accuracy and inference speed. GO-SLAM \cite{zhang2023goslam} introduces an effective loop closure strategy that enables global correction but may impact tracking speed and efficiency. However, these methods result in severe errors in reconstruction and tracking in dynamic scenes due to interference from high/low dynamic objects. That leads to incorrect data association and cumulative drift amplification in pose estimation. The dynamic objects as interference information affecting the reconstruction results, causing artifacts or catastrophic forgetting issues. Although NICE-SLAM \cite{zhu2022nice} proposes a pixel filtering method based on depth threshold to achieve effective tracking and reconstruction in low dynamic scenes, its inference speed and tracking accuracy decrease significantly in high-dynamic(involving significant human movement and high speed) scenes with occlusion and interference. Meanwhile, in challenging scenarios \cite{shi2020openloris}, neural implicit SLAM systems often get weak in pose estimation, resulting in tracking loss or significant pose drift. Neural implicit SLAM methods usually cannot match traditional methods in tracking accuracy due to the lack of loop detection and global BA.
We combine the advantages of traditional methods and neural implicit methods, using semantic constraints to limit the impact of dynamic objects during tracking. We segment the dynamic, static, and potentially dynamic objects to achieve adaptive background restoration. In the rendering process, we also utilize semantic information as an additional constraint loss to restrict high-dynamic occlusion and artifacts. We are able to adaptively remove high-dynamic objects during the rendering process and retain potential dynamic objects to enhance the completeness of scene reconstruction.

\begin{figure*}[!b]
  \centering
  \includegraphics[width=\textwidth,height=15cm,keepaspectratio]{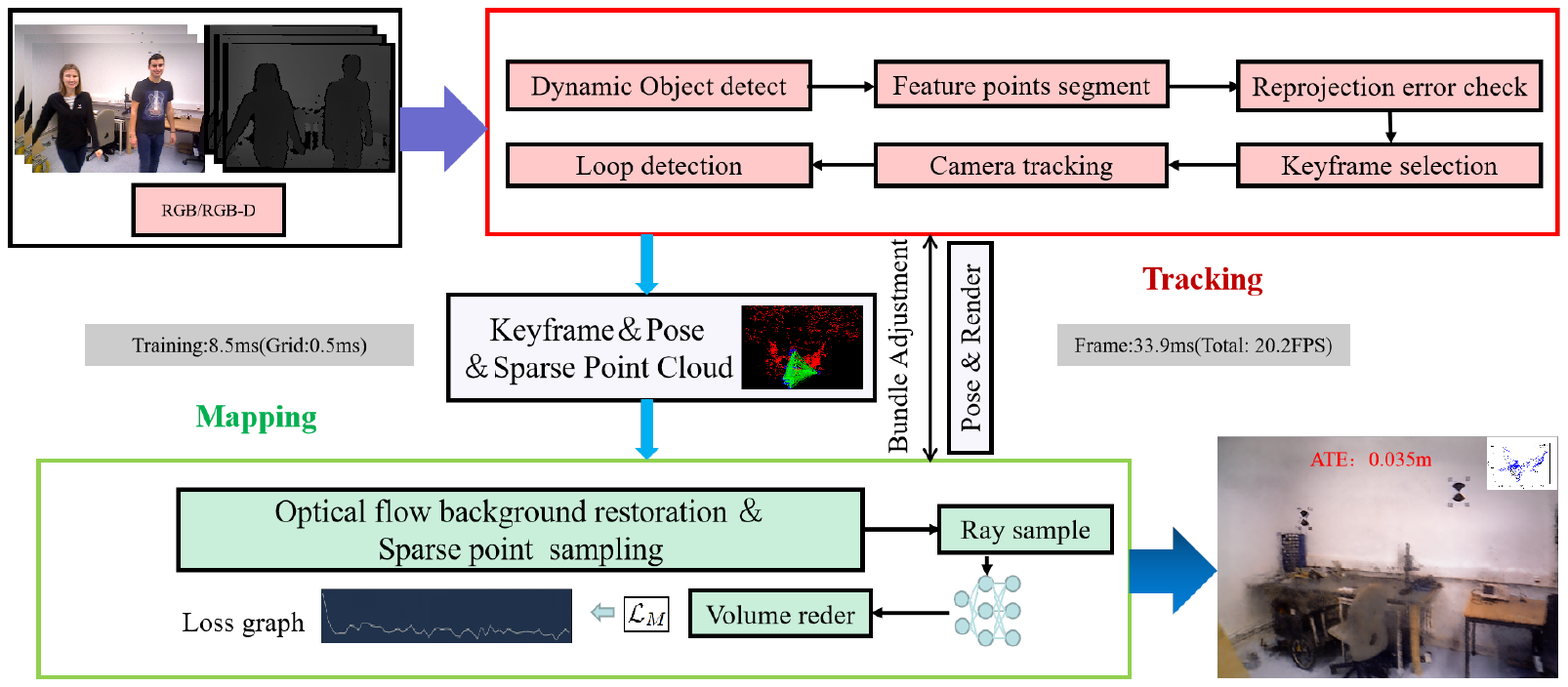}
  \caption{\textbf{System Overview.} Our DDN-SLAM system consists of two main modules: tracking and mapping, divided into four threads. Our four threads optimize alternately. The segmentation thread detects and segments dynamic feature points and pixels, suppressing potential feature points. The tracking thread extracts feature points, receives feature points filtered by conditional filtering for tracking, obtains static optical flow, generates keyframes and camera poses. The mapping thread receives background segmentation masks for high and low dynamics, performs keyframe generation and volume rendering. The loop detection thread detects loops and performs global bundle adjustment. Our system can be updated in real time.}
  \label{fig:fig3}
\end{figure*}

\section{Method}
Our system framework is shown in Figure 1. We introduce our method in three parts: Sec.\ref{3.1} presents our foreground segmentation technique based on semantic understanding and Gaussian distribution assumptions. We employ feature points that exclude dynamic disturbances for tracking and utilize a secondary verification of reprojection errors to re-establish long-term data associations.  Sec.\ref{3.2} introduce our mixed method to background restoration, where we utilize sparse static optical flow to segment low-dynamic pixels and employ background inpainting to address occlusion issues caused by dynamic objects. We enhance the quality of static reconstruction by incorporating jump sampling guided by sparse feature points. In Sec. \ref{3.3}, we introduced our mapping method where we represent the scene using NeRF. We proposed a dynamic loss based on motion consistency and the color and depth of dynamic pixels, which can effectively reduce the impact of highly dynamic pixels.

\subsection{Foreground and Background Segmentation}\label{3.1}

Based on ORBSLAM3\cite{campos2021orb}, we have constructed a feature point tracking system. Unlike other neural implicit SLAM systems, to address the interference introduced by dynamic objects in the real world, we incorporate prior semantic information through YOLOv9\cite{yolov9github} to obtain dynamic bounding box priors and form a set of dynamic points. To segment the dynamic feature points within the bounding box in a limited manner, we introduce a background hypothesis based on a Gaussian mixture distribution. Given that the bounding boxes constrained by semantic information contain dynamic objects, we set the collection of depth values for all pixels as $D_n = \{d_1, d_2, \ldots, d_N\}$. We define the average depth value of the four corners of the Yolo box as $\bar{d}_{\text{corner}}$, and we use two Gaussian distributions, $\mathcal{N}(\mu_{\text{fg}}, \sigma_{\text{fg}}^2)$ and $\mathcal{N}(\mu_{\text{bg}}, \sigma_{\text{bg}}^2)$, to represent the depth value distributions of the foreground and background, respectively, where the subscripts fg and bg denote foreground and background. We introduce prior probabilities, assuming pixels with depth greater than $\bar{d}_{\text{corner}}$ as background, and those with lesser depth as foreground. For depth values greater than $\bar{d}_{\text{corner}}$, the background prior probability $\pi_{\text{bg}}(d_i) = 0.9$ and the foreground prior probability $\pi_{\text{fg}}(d_i) = 0.1$.

The probability density function (PDF) of the Gaussian Mixture Model(GMM) for a depth value \(d_i\) is defined as:
\begin{equation}
p(d_i) = \pi_{\text{fg}}(d_i) \cdot \mathcal{N}(d_i | \mu_{\text{fg}}, \sigma_{\text{fg}}^2) + \pi_{\text{bg}}(d_i) \cdot \mathcal{N}(d_i | \mu_{\text{bg}}, \sigma_{\text{bg}}^2)
\end{equation}
where \(\pi_{\text{fg}}\) and \(\pi_{\text{bg}}\) are the mixing coefficients, satisfying \(\pi_{\text{fg}} + \pi_{\text{bg}} = 1\).

Based on the current parameter estimates, we calculate the expected value of the latent variables or the expected distribution. We obtain the posterior probability of each depth value belonging to the foreground or background, denoted as \(\gamma_{\text{fg}}(d_i)\) and \(\gamma_{\text{bg}}(d_i)\), respectively:
\begin{equation}
\gamma_{\text{fg}}(d_i) = \frac{\pi_{\text{fg}}(d_i) \cdot \mathcal{N}(d_i | \mu_{\text{fg}}, \sigma_{\text{fg}}^2)}{p(d_i)}
\end{equation}
\begin{equation}
 \gamma_{\text{bg}}(d_i) = \frac{\pi_{\text{bg}}(d_i) \cdot \mathcal{N}(d_i | \mu_{\text{bg}}, \sigma_{\text{bg}}^2)}{p(d_i)}
\end{equation}
\begin{figure}[htbp]
  \centering
  \includegraphics[width=0.5\linewidth]{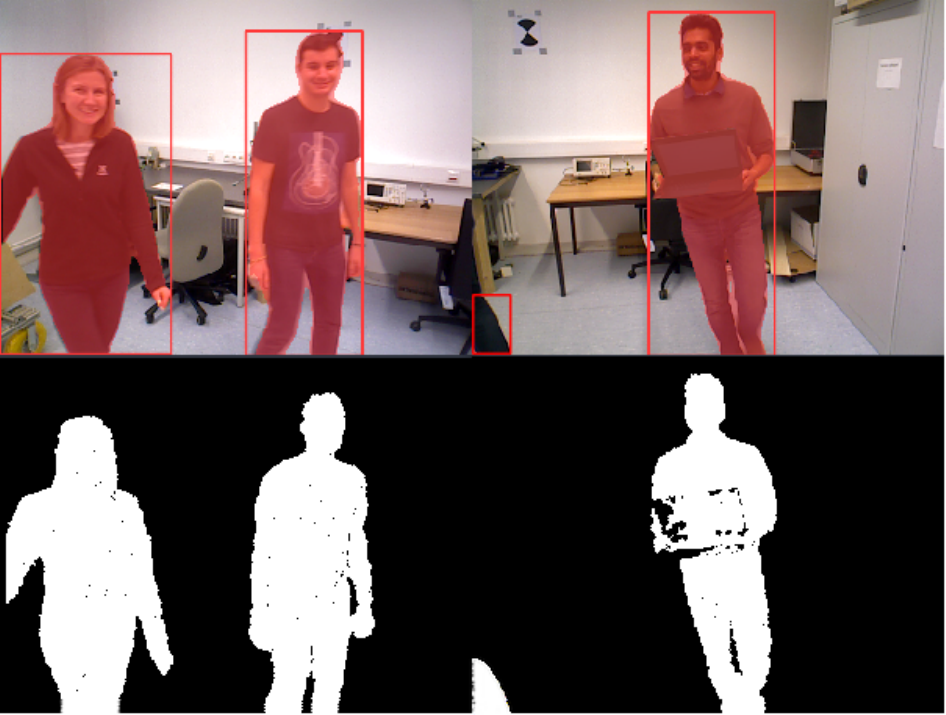}
  \captionsetup{justification=justified,singlelinecheck=false}
  \caption{Our pixel-level clustering results of foreground depth probability verification demonstrate the accurate segmentation of dynamic object masks. The top section shows the segmentation results of YOLOv9\cite{yolov9github}, while the bottom section displays our clustering results.}
  \label{fig:fig5}
\end{figure}

Parameters are updated based on the expected values to maximize the likelihood function. We update the model parameters, including the means, variances, and mixing coefficients of the two Gaussian distributions:
\begin{equation}
\mu_{\text{fg}} = \frac{\sum_{i=1}^{N} \gamma_{\text{fg}}(d_i) \cdot d_i}{\sum_{i=1}^{N} \gamma_{\text{fg}}(d_i)}\ , \  \sigma_{\text{fg}}^2 = \frac{\sum_{i=1}^{N} \gamma_{\text{fg}}(d_i) \cdot (d_i - \mu_{\text{fg}})^2}{\sum_{i=1}^{N} \gamma_{\text{fg}}(d_i)} 
\end{equation}
\begin{equation}
 \mu_{\text{bg}} = \frac{\sum_{i=1}^{N} \gamma_{\text{bg}}(d_i) \cdot d_i}{\sum_{i=1}^{N} \gamma_{\text{bg}}(d_i)}\ , \  \sigma_{\text{bg}}^2 = \frac{\sum_{i=1}^{N} \gamma_{\text{bg}}(d_i) \cdot (d_i - \mu_{\text{bg}})^2}{\sum_{i=1}^{N} \gamma_{\text{bg}}(d_i)} 
\end{equation}

Based on the convergence results of the EM (Expectation-Maximization) algorithm, we can assign the most likely distribution to each pixel. If \(\gamma_{\text{fg}}(d_i) > \gamma_{\text{bg}}(d_i)\), then \(d_i\) is classified as foreground; otherwise, it is considered background. Thus, we can obtain the set of static feature points \(P_{\text{static}} = \{p_1, p_2, \ldots, p_N\}\) and the set of dynamic feature points \(P_{\text{dynamic}} = \{p_1, p_2, \ldots, p_N\}\). If more than 10\% of points within a box are dynamic, we consider the bounding box to contain a dynamic object. This ensures the consistency of object motion in the scene. The depth segmentation results after segmenting foreground and background on the Bonn\cite{palazzolo2019refusion} dataset are shown in Figure 2.

To avoid excluding potential static points, we introduce the mean reprojection error $e_m(x)$ as a check before bundle adjustment (BA). Consequently, we assign weights $w_i$ to static and dynamic points as follows:
\begin{equation}
w_i = \gamma_{\text{bg}}(d_i)
\end{equation}
Where, $w_i$ is directly assigned based on the posterior probability $\gamma_{\text{bg}}(d_i)$ of point $i$ belonging to the background, ensuring that the weights reflect the likelihood of each point being a static feature point.

\begin{figure}
  \centering
  \includegraphics[width=0.48\textwidth]{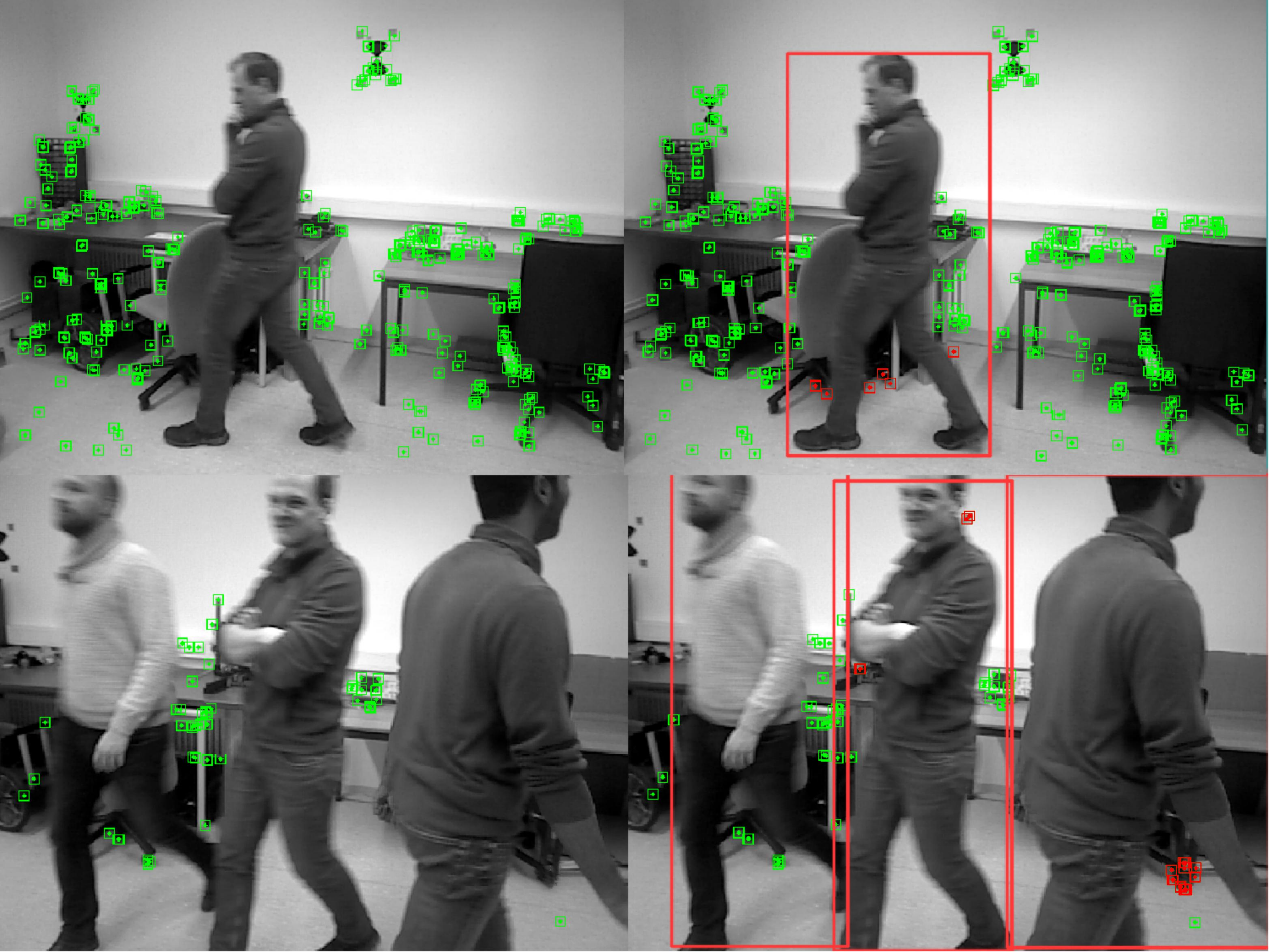}
  \caption{Our method achieves accurate segmentation of multiple targets in high-dynamic scenes of the Bonn dataset \cite{palazzolo2019refusion} with multiple targets, where the human body is framed as a semantic framework for detection. We remove feature points within dynamic human bodies and preserve static feature points within the bounding box. Green marks the preserved feature points, and the red markers indicate the recovered feature points.}
  \label{fig:fig2}
\end{figure}

The reprojection error based on features classified as static, whose positions are more likely to represent the camera's true movement relative to the environment. Specifically, we disable the function of eliminating potential feature points in static scenes. Therefore, we define the translation vector optimization process in pose estimation as follows:
\begin{equation}
e_i(x) = w_i|x_i - \pi(R X_i + T_{\text{static}})|
\end{equation}
\begin{equation}
e_m(x) = \frac{1}{N_{\text{static}}} \sum_{k=1}^{N_{\text{static}}} |x - \pi(R_k X + T_{\text{static},k})|
\end{equation}
where $\pi(\cdot)$ is the projection transformation from the camera coordinate system to the plane coordinate system, and $R$ and $R_k$ are the rotation matrix vector of the $k_{th}$ frame. $T_{\text{static}}$ and $T_{\text{static},k}$ represent the translation vectors of the $k_{th}$ frame adjusted based on the static point set.

In pose estimation, the optimization process should focus more on static feature points. We use the translation vector $T_{\text{static}}$, derived from static feature points, to minimize the reprojection error. We jointly optimize the rotation matrix $R$, the translation vector $T_e$, the mapping of 3D map points to image coordinates $P_e$, and the weights $w_i$ of each map point. We obtain the minimum reprojection error $L_{rp}$ as follows:
\begin{equation}
\min_{\left\{[R | T_{\text{static}}]\right\},\left\{P_e\right\}} L_{rp} = \underset{\left\{[R | T_{\text{static}}]\right\},\left\{P_e\right\}}{\operatorname{argmin}} \sum_{i=0}^{N_{\text{static}}} \rho\left(e_i(x)^T \mathbf{\Theta} e_i(x)\right)
\end{equation}
where $\mathbf{\Theta}$ is the information matrix related to the scale of the feature points, $\rho$ is the robust Huber loss function. We performed BA to optimize the keyframes and camera poses as well as the observed map points in these keyframes.  Foreground points within dynamic boxes are considered as dynamic points and excluded during pose initialization. The optimized keyframes and mapped points were then passed to NeRF for further processing. The feature point segmentation results are shown in Figure 3.

\subsection{Mixed Background Restoration}\label{3.2}

To eliminate ghosting artifacts caused by dynamic occlusions while preserving potential dynamic objects, we optimize both the input rendering and the rendering process. We propose a mixed background restoration strategy. This allows us to reduce dynamic interference and enhance static reconstruction.

\noindent\textbf{Optical flow-based background restoration}: Inspired by \cite{zhang2020flowfusion}, \cite{bescos2018dynaslam}, and \cite{ye2022deflowslam}, we combine optical flow-based dynamic judgment with background completion for background restoration. Different from traditional dynamic SLAM methods, which require complete traversal and completion of all input frames, resulting in extremely high computational costs. We only need to perform optical flow-based dynamic pixel segmentation for the keyframes involved in NeRF mapping, dividing the image into high-dynamic, low-dynamic, and static regions. We then fill in the low-dynamic pixels, thereby improving sampling efficiency and reducing memory consumption. We retain static pixels that do not interfere with the tracking process to ensure that our method can accurately distinguish dynamic and static information without interference. We retain static pixels that do not interfere with the tracking process to ensure that our method can correctly distinguish dynamic and static information without interference. We use the pixel mask $M_d$ obtained from the Gaussian mixture distribution clustering within the bounding box in Sec. \ref{3.1}, as well as the filtered static feature points. We construct the sparse optical flow vector $V_{s}^{\text{static}}$ between adjacent frames as follows. 
\begin{equation}
V_{s}^{\text{static}} = V_{s}^{\text{scene}} - V_{\text{observed}}
\end{equation}
Where $V_{s}^{\text{static}}$ refers to the static optical flow, which is the difference between the scene-induced optical flow $V_{s}^{\text{scene}}$ and the observed optical flow $V_{\text{observed}}$ caused by camera motion. Based on temporal consistency, we calculate the dynamic optical flow vector $M_{si}^{\text{dynamic}}$ for each pixel in $M_d$ between two consecutive frames.
In the mapping process, the high-dynamic parts of objects are removed due to the constraints we imposed in Sec.\ref{3.3}. Therefore, we only need to restoration the low-dynamic pixels to avoid interference caused by them. We can further refine $M_d$ by dividing it into low-dynamic and high-dynamic pixels to reduce its occupancy. We establish a threshold $\delta$ that defines pixels with optical flow. We obtain the low-dynamic pixel optical flow set $F_{dl}$, high-dynamic pixel optical flow set $F_{dh}$, and static pixel optical flow set $F_{ds}$, with the partition formula as follows:
\begin{equation}
\text{Pixel Type} = \begin{cases} 
F_{ds}, & \text{if } V \leq V_{s}^{\text{static}} \\
F_{dl}, & \text{if } V_{s}^{\text{static}} < V < \delta \\
F_{dh}, & \text{if } V \geq \delta
\end{cases}
\end{equation}

Then, we establish a sliding window that bundles the latest 20 keyframes into a keyframe group and calculate the relative poses between each keyframe and the current frame. We fill in the color image and depth map corresponding to the low dynamic part mask $M_{dl}$. For RGB mode, we fill in the depth map obtained from\cite{bhat2023zoedepth}. We use the repaired keyframe image as the final mapping frame.

\noindent\textbf{Sparse point cloud-guided skip sampling}: Inspired by \cite{chung2023orbeezslam}, we input the static sparse point cloud acquired by our system into the backend NeRF mapping thread, considering the corresponding map points of the sparse point cloud. We obtain the spatial position of map points in the density grid through their poses. Given the correlation between map points and feature points, the surroundings of the map points are closer to the surface of static objects. Therefore, we apply sampling adjustment to the filtered static point cloud. We use the sample counter of the density network to calculate the relationship between the sampling rays and the map points. When a voxel in the density grid intersects with a map point, we mark that voxel as 1; otherwise, it is marked as 0. Each voxel marked as 1 indicates that the corresponding voxel should have an additional 10 ray samples. We set the upper limit of the sampling threshold as $\theta$. This method allows for a more complete reconstruction of static surfaces, increasing the proportion of static pixels and better filling in the gaps.

\subsection{Dynamic Volume Rendering}\label{3.3}

 For representing 3D scenes in the presence of occlusion, we employ a volumetric rendering process combined with multi-resolution hash encoding \cite{muller2022instant}. The given position $x$ can be mapped to the respective positions on each resolution grid, and trilinear interpolation is performed on the grid cell corners to obtain concatenated encoding feature vectors $\boldsymbol{\gamma}_t(\mathbf{x}_t,\mathbf{t}) \in \mathbb{R}^c$ . After feature concatenation, we input them into an MLP.
More specifically, our SDF network $F_{\Phi_{sdf}}$ is composed of a single MLP, with learnable parameters $\Phi_{sdf}$. It takes the point position $x$ and the corresponding hash encoding $h_{\Phi_{hash}}(\mathbf{x})$ as inputs and predicts the signed distance field (SDF) as:

\begin{equation}
\Phi_{sdf} , \mathbf{t}=F_{\Phi_{sdf}}(\mathbf{x}, h _{\Phi_{hash}}(\mathbf{x})).
\end{equation}

To estimate the color $\mathbf\Phi(\mathbf{x})$, we utilize a color network $F_{\Phi_{\text{color}}}$ that processes the learned geometric feature vector $\mathbf{t}$, position $\mathbf{x}$, and SDF gradient $\mathbf{n}$ with respect to $\mathbf{x}$:
\begin{equation}
\mathbf{\Phi}(\mathbf{x}) = {F_{\Phi_{\text{color}}}(\boldsymbol{\gamma}(\mathbf{x}), {t})}.
\end{equation}

We adopt hierarchical sampling to obtain $N_{\text{strat}}$ samples on each ray, and additionally sample $N_{\text{imp}}$ points near the surface. For all $N = N_{\text{strat}} + N_{\text{imp}}$ points on the ray. We sampled N and denoted as $x_k=\mathbf{o}+t_k\mathbf{d}$. This process represents the projection of a ray $\mathbf{r}$ from the camera center $\mathbf{o}$ through a pixel along its normalized viewing direction $\mathbf{d}$. We adopt the SDF representation and express its opacity as $\alpha_i$.
\begin{equation}
\alpha_k=\max \left(\frac{\Omega\left(\Phi_s\left(\mathbf{x}_k\right)\right)-\Omega\left(\Phi_s\left(\mathbf{x}_{k+1}\right)\right)}{\Omega\left(\Phi_s\left(\mathbf{x}_k\right)\right)}, 0\right)
\end{equation}

where $\Omega$ is the sigmoid function. $T_k = \prod_{j=1}^{k-1}\left(1-\alpha_j\right)$ denotes the cumulative transmission rate, which represents the proportion of light reaching the camera. With the weight ${T_k}$ We use MLP predictions to render the color and depth for each ray:

\begin{equation}
\hat{\mathbf{C}}=\sum_{k=1}^N {T_k} {\alpha_k} \mathbf{\Phi}(\mathbf{x})\quad       \hat{\mathbf{D}}=\sum_{k=1}^N {T_k} {\alpha_k} {t_k}
\end{equation}

The definitions of color and depth rendering losses are as follows:

\begin{equation}
\mathcal{L}_{\text{rgb}} = \frac{1}{N} \sum_{n=1}^{N} \left( \hat{\mathbf{C}}_n - \mathbf{C}_n \right)^2, \quad \mathcal{L}_{\text{depth}} = \frac{1}{N} \sum_{n=1}^{N} \left( \hat{\mathbf{D}}_r - \mathbf{D}_r \right)^2
\end{equation}

where \(\hat{\mathbf{D}}_r\) represents the set of rays with valid depths, \(\mathbf{D}_r\) corresponds to the filtered set of pixel depths, and \(R_s\) is the batch of rays contributing to depth. In monocular mode, supervision is performed using predicted depths from \cite{bhat2023zoedepth}. 
To effectively reduce the impact of dynamic pixels and improve the accuracy of depth and color predictions, we construct a dynamic rendering loss. Our loss function consists of three parts: the basic loss (including depth loss and color loss), the motion consistency loss, the dynamic area penalty loss (\(\mathcal{L}_{\text{pl}}\)). The basic loss evaluates the overall accuracy of the prediction, while the additional depth and color penalty items are specifically aimed at dynamic pixel areas to constrain the occlusion impact of dynamic objects. We construct the motion consistency loss (\(\mathcal{L}_{\text{m}}\)) as follows:
\begin{equation}
\mathcal{L}_{\text{m}} = \frac{1}{N_{\text{dynamic}}} \sum_{i \in \text{Dynamic}} \left\| R_{d_i}^{\text{pred}} - R_{d_i}^{\text{actual}} \right\|^2
\end{equation}
where \(R_{d_i}^{\text{pred}}\) is the predicted motion distance for the \(i\)th dynamic pixel, \(R_{d_i}^{\text{actual}}\) is its actual motion distance, and \(N_{\text{dynamic}}\) is the number of dynamic pixels.
 We construct the dynamic area penalty item (\(\mathcal{L}_{\text{pl}}\)) as:
\begin{equation}
\mathcal{L}_{\text{pld}} = \frac{1}{N_{\text{dynamic}}} \sum_{i \in \text{Dynamic}} \left\| \hat{\mathbf{D}}_i - \mathbf{D}_i \right\|^2, \quad \mathcal{L}_{\text{plc}} = \frac{1}{N_{\text{dynamic}}} \sum_{i \in \text{Dynamic}} \left\| \hat{\mathbf{C}}_i - \mathbf{C}_i \right\|^2
\end{equation}
where \(N_{\text{dynamic}}\) is the number of pixels in the dynamic area, \(\hat{\mathbf{D}}_i\) and \(\mathbf{D}_i\) respectively represent the predicted and actual depth values. Combining the above losses, we define the rendering loss as:
\begin{equation}
\mathcal{L}_{\text{R}} = \lambda_r \mathcal{L}_{\text{rgb}} + \lambda_d \mathcal{L}_{\text{depth}} + \lambda_{\text{dynamic}} (\mathcal{L}_{\text{pld}} + \mathcal{L}_{\text{plc}}) + \lambda_{\text{m}} \mathcal{L}_{\text{m}}
\end{equation}
where \(\lambda_{\text{dynamic}}\) and \(\lambda_{\text{m}}\) are initialized to 0.001.

\section{Experiment}

\subsection{Experimental Setup}
\textbf{Implementation Details. }Our system is implemented on a system with a NVIDIA RTX 3090ti GPU and an Intel Core i7-12700K CPU operating at 3.60 GHz. We use the YOLOv9\cite{yolov9github} network pre-trained on the COCO dataset\cite{lin2014microsoftcoco} as the backbone of our object detection system. 
We initialize the optical flow threshold $\delta$ to 5, and the sampling threshold $\theta$ to 500. Due to the page limit the more parameter settings and detailed ablation experiments can be found in the supplementary materials.

\noindent\textbf{Dataset. }We evaluated our system on a total of 2 dynamic datasets, 1 challenging dataset with large viewpoints and weak textures, and 3 static datasets. The datasets used for evaluation include TUM RGB-D \cite{sturm2012evaluating}, Bonn \cite{palazzolo2019refusion}, OpenLORIS-Scene \cite{shi2020openloris}, Replica\cite{straub2019replica}, ScanNet\cite{dai2017scannet}, and EuRoC \cite{Burri2016}. The TUM RGB-D \cite{sturm2012evaluating} dataset consists of 6 high-dynamic sequences, 2 low-dynamic sequences, and 3 static sequences. The Bonn \cite{palazzolo2019refusion} dataset includes 6 high-dynamic range sequences and 2 sequences with complex occlusions. The challenging scenes dataset consists of 7 indoor sequences from the OpenLORIS-Scene dataset \cite{shi2020openloris}, demonstrating large changes in viewpoint, variations in lighting, and sudden dynamic interferences. The static datasets consist of 6 sequences from the ScanNet \cite{dai2017scannet} dataset, 8 sequences from the high-quality virtual scene dataset Replica\cite{straub2019replica}, and 3 Sequences on the EuRoC \cite{Burri2016} Dataset. Due to the page limit, we include more experimental results in the supplemental material.

\noindent\textbf{Metrics and Baseline. }We evaluate the reconstruction quality using \{Depth L1\} (cm), \{Accuracy\} (cm), \{Completion\} (cm), \{Completion Ratio\} (\%), and \{PSNR\} (dB). For assessing camera tracking quality in dynamic scenes, we use \{ATE RMSE\}. Our main baselines for reconstruction quality and camera tracking include iMAP \cite{sucar2021imap}, NICE-SLAM \cite{zhu2022nice}, ESLAM \cite{bloesch2018codeslam}, GO-SLAM \cite{zhang2023goslam}, Orbeez-SLAM\cite{chung2023orbeezslam}and Co-SLAM \cite{wang2023coslam}. We utilized the replication code provided by \cite{kong2023vmap} to achieve better performance reproduction for iMAP \cite{sucar2021imap}. For a fair comparison, we introduce classical traditional dynamic SLAM methods for tracking results in dynamic scenes, including LC-CRF SLAM \cite{du2020accurate}, Crowd-SLAM \cite{soares2021crowdslam}, ElasticFusion \cite{whelan2015elasticfusion}, ORB-SLAM2 \cite{mur2017orb}, and DS-SLAM \cite{yu2018ds}.

\subsection{Evaluation in Dynamic Scenes}

\begin{figure*}[!t]
  \centering
  \makebox[\textwidth]{\includegraphics[width=0.8\paperwidth]{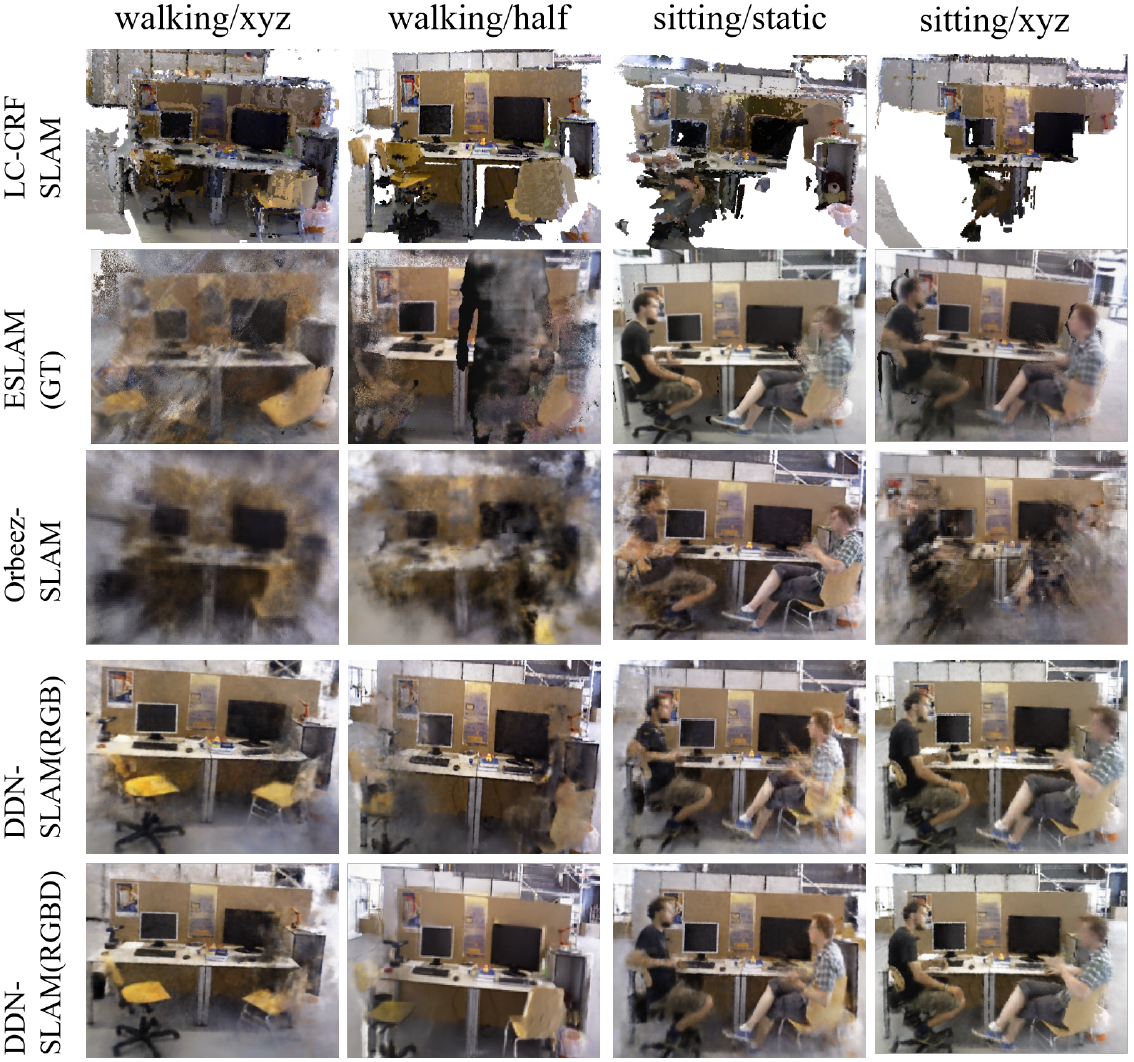}}
  \captionsetup{justification=justified,singlelinecheck=false}
  \caption{We compared the reconstruction results of traditional LC-CRF\cite{du2020accurate} SLAM, ESLAM\cite{yang2019cubeslam}, Orbeez-SLAM\cite{chung2023orbeezslam}, and our method (in both RGB and RGB-D modes) on dynamic sequences from TUM RGB-D \cite{sturm2012evaluating}. We presented results on four TUM dynamic sequences, with two being high-dynamic sequences (involving significant human movement and high speed) and two being low-dynamic sequences (involving minimal human movement and slower speed). Our method demonstrates our ability to preserve potential dynamic objects, eliminate occlusion interference, and reduce the occurrence of artifacts, achieving high-quality reconstruction. Specifically, for a fair comparison, we employed rendering results based on ground truth poses for ESLAM to avoid reconstruction errors caused by tracking inaccuracies.}
  \label{fig:fig4}
\end{figure*}

\noindent\textbf{TUM RGB-D \cite{sturm2012evaluating} Dynamic Sequence Evaluation. }To evaluate our system on TUM RGB-D \cite{sturm2012evaluating} dynamic sequences, we compare it with LC-CRF SLAM \cite{du2020accurate}, Crowd-SLAM \cite{soares2021crowdslam}, ORB-SLAM2 \cite{mur2017orb}, NICE-SLAM \cite{zhu2022nice}, ESLAM \cite{yang2019cubeslam} and Co-SLAM \cite{wang2023coslam}. Our method achieves effective rendering, effectively avoids occlusion interference caused by high-dynamic human subjects, reasonably fills scene holes, and achieves high-precision geometric detail reconstruction with minimal artifacts. Our method preserves the potential dynamic objects and enhances the integrity of the reconstruction. as shown in Figure 4. We have achieved tracking results that are competitive with traditional methods, as shown in Table 2. Due to severe accumulated drift, current neural implicit SLAM methods are unable to complete reliable mapping.

\begin{table*}[!t]
\begin{threeparttable}
\renewcommand\arraystretch{1.3} 
\resizebox{\textwidth}{!}{
\begin{tabular}{c|cccccccccc}
\hline
                                                                                         &   Completion Rate                         & walking/xyz & walking/half & walking/static & walking/rpy & sitting-xyz & sitting-half & AVG  \\ \hline
\multirow{1}{*}{LC-CRF SLAM \cite{du2020accurate}}                                                         &\textbf{ 100\% }                     & 0.020 & \underline{0.030} & {0.028} & 0.051 & 0.013 & 0.026 &0.028     \\ 
\multirow{1}{*}{Crowd-SLAM \cite{soares2021crowdslam}}                                                         &\textbf{ 100\%}                     & \underline{0.019} & 0.036 &  \textbf{0.007} & \underline{0.044} & 0.017 & 0.025  &\underline{0.027}   \\ 
\multirow{1}{*}{ORB-SLAM2 \cite{mur2017orb}}                                                             & \underline{93\% }                & 0.721 & 0.462 & 0.386 & 0.785 & \textbf{0.009} & 0.026  &0.373  \\ 
\hline
\multirow{1}{*}{NICE-SLAM \cite{zhu2022nice}}                                                           & 79\%                     & 0.420 & 0.732 & 0.491 & N/A & 0.029 & 0.134  &0.362  \\ 
\multirow{1}{*}{ESLAM \cite{bloesch2018codeslam}}                                                               & 61\%                  & 0.432 & N/A & 0.075  & N/A & 0.034 &0.063   &0.151\\  
\multirow{1}{*}{\begin{tabular}[c]{@{}l@{}}Co-SLAM \cite{wang2023coslam}\end{tabular}}                & 44\%                   & 0.714    & N/A    & 0.499    &N/A  & 0.065  &\underline{0.023}   & 0.325 \\        
\hline
\multirow{1}{*}{DDN-SLAM(RGB)}                                                           & \textbf{100\%}                     & {0.028}    &{0.041}    & \textbf{0.025}    & {0.089} & {0.013}  & {0.031}   & {0.037}\\
\multirow{1}{*}{DDN-SLAM}                                                           &\textbf{ 100\% }                    & \textbf{0.014}    &\textbf{0.023}    & \underline{0.010}    & \textbf{0.039} & \underline{0.010}  & \textbf{0.017}   & \textbf{0.020}\\\hline
\end{tabular}}
\end{threeparttable} 
\noindent\caption {{The tracking results on 8 dynamic sequences in TUM RGB-D \cite{sturm2012evaluating}(ATE[m]).  We also present the average tracking Completion Rate of 6 scenes. The best results are indicated in bold, and the second-best results are indicated with a horizontal line. N/A indicates failure in all 5 measurements and the total number of frames with successful tracking is less than 50\% of the total.  }}\label{tab1}      
\end{table*}

\noindent\textbf{Bonn \cite{palazzolo2019refusion} Dynamic Sequence Evaluation. } Our tracking method demonstrates robustness and achieves competitive tracking results, as demonstrated in Table 3. Compared to the TUM RGB-D \cite{sturm2012evaluating} dataset, the Bonn \cite{palazzolo2019refusion} dataset exhibits more complex dynamic scenes, severe occlusions, and the presence of multiple moving objects simultaneously. The reconstruction results are shown in Fig. 5. In comparison to traditional dynamic SLAM methods, our approach achieves better results. The current neural implicit systems have shown significant drift or tracking failures in most scenarios.

\begin{table*}[!t]
\begin{threeparttable}
\renewcommand\arraystretch{1.3} 
\resizebox{\textwidth}{!}{
\begin{tabular}{c|cccccccccccc}
\hline
                                                                                         & Completion Rate                    & balloon   & balloon2   & move & move2 & crowd & crowd2  & person & person2 & AVG  \\ \hline
\multirow{1}{*}{LC-CRF SLAM \cite{du2020accurate}}                                                         &\textbf{ 100\%}            & \underline{0.027}     &0.024   & 0.079 & 0.186 & 0.027 & 0.098 &\underline{0.046} & \underline{0.041} &0.066     \\ 
\multirow{1}{*}{Crowd-SLAM \cite{soares2021crowdslam}}                                                         &\textbf{ 100\%}             & 0.037     &\textbf{0.014}         & \underline{0.026} & \textbf{0.029} & \underline{0.019} & \underline{0.035} & 0.046 & 0.099  &\underline{0.038}   \\ 
\multirow{1}{*}{ORB-SLAM2 \cite{mur2017orb}}                                                             &\underline{ 91\% }        & 0.065     &0.230         & 0.320 & 0.039 & 0.496 & 0.989 & 0.692 & 0.079  &0.363  \\ 
\hline
\multirow{1}{*}{NICE-SLAM \cite{zhu2022nice}}                                                           & 87\%                     &2.442  &2.018     &0.177  &0.832 &1.934 &3.582 &0.245 &0.536 &1.470   \\ 
\multirow{1}{*}{ESLAM \cite{bloesch2018codeslam}}                                                               & 64\%           &0.203  &0.235      &0.190    &0.129  &0.416    &1.142    & 0.845  & 7.441  &1.325  \\ 
\multirow{1}{*}{\begin{tabular}[c]{@{}l@{}}Co-SLAM \cite{wang2023coslam}\end{tabular}}                & \textbf{100\%}              &0.211  &0.480     & 0.076 &0.200 & 51.983 & 2.019 &0.695 & 0.758  &7.052   \\     
\hline    
\multirow{1}{*}{DDN-SLAM}                                                           & \textbf{100\% }           & \textbf{0.018} &  \underline{0.041}   & \textbf{0.020} &  \underline{0.032}  &\textbf{0.018}    & \textbf{0.023}    & \textbf{0.043}    & \textbf{0.038} &\textbf{0.029}     \\\hline
\end{tabular}}
\end{threeparttable} 
\caption {{Tracking results on 8 dynamic sequences in the Bonn\cite{palazzolo2019refusion} dataset(ATE[m]).  We also present the average tracking Completion Rate of 8 scenes. The best results are indicated in bold, and the second-best results are indicated with a horizontal line. For methods that fail to track, we report the best result obtained before the failure.}}\label{tab1}      
\end{table*}

\subsection{Evaluation in Challenging Scenes}

\noindent\textbf{Openloris-Scene \cite{shi2020openloris} Dataset Experimental Results. }We conducted reconstruction and tracking evaluation on the challenging Openloris-Scene \cite{shi2020openloris} dataset. Our method significantly outperforms existing neural implicit methods and traditional approaches. Benefite from accurate feature point segmentation, and there is no significant drift as in NICE-SLAM \cite{zhu2022nice} and Co-SLAM \cite{wang2023coslam}, as shown in Table 5. Our method is capable of tracking in highly challenging scenarios, and improving tracking completion rates.
\begin{table*}[h]
\begin{threeparttable}
\renewcommand\arraystretch{1.3} 
\resizebox{\textwidth}{!}{
\begin{tabular}{c|cccccccccc}
\hline
& Completion Rate & office1-1 & office1-2 & office1-3 & office1-4 & office1-5 & office1-6 & office1-7 & AVG \\ \hline
\multirow{1}{*}{ORB-SLAM2 \cite{mur2017orb}} & \cellcolor{red!25}100\%↑ & \cellcolor{green!25}0.074 & \cellcolor{green!25}0.072 & \cellcolor{yellow!25}0.015 & \cellcolor{yellow!25}0.077 &0.161 & \cellcolor{green!25}0.080 & \cellcolor{red!25}0.100 & \cellcolor{yellow!25}0.082 \\
\multirow{1}{*}{ElasticFusion \cite{whelan2015elasticfusion}} &\cellcolor{red!25}100\%↑ & 0.112 & 0.118 & 0.022 & 0.100 & 0.144 & 0.113 & 0.192 & 0.114 \\
\multirow{1}{*}{DS-SLAM \cite{yu2018ds}} & \cellcolor{yellow!25}77\%↑ & \cellcolor{yellow!25}0.089 & \cellcolor{yellow!25}0.101 & 0.016 & \cellcolor{red!25}0.021 & \cellcolor{red!25}0.052 & 0.129 & \cellcolor{green!25}0.077 & \cellcolor{green!25}0.069 \\
\hline
\multirow{1}{*}{NICE-SLAM \cite{zhu2022nice}} & \cellcolor{red!25}100\%↑ & 0.110 & 0.142 & 0.040 & 0.232 & \cellcolor{yellow!25}0.130 & 0.104 & 0.251 & 0.144 \\
\multirow{1}{*}{ESLAM \cite{bloesch2018codeslam}} & \cellcolor{red!25}100\%↑ & 0.101 & 0.152 & \cellcolor{red!25}0.010 & 0.235 & 0.151 & 0.101 & 0.201 & 0.150 \\
\multirow{1}{*}{\begin{tabular}[c]{@{}l@{}}Co-SLAM\cite{wang2023coslam}\end{tabular}} & \cellcolor{green!25}99\%↑ & 0.112 & 0.173 & 0.043 & 0.141 & 0.174 & 0.134 & 0.186 & 0.138 \\
\multirow{1}{*}{GO-SLAM \cite{zhang2023goslam}} & \cellcolor{red!25}100\%↑ & 0.120 & 0.160 & 0.033 & 0.183 &  0.942 & \cellcolor{yellow!25}0.092 & 0.142 & 0.239 \\
\hline
\multirow{1}{*}{DDN-SLAM} & \cellcolor{red!25}100\%↑ & \cellcolor{red!25}0.058 & \cellcolor{red!25}0.071 & \cellcolor{green!25}0.011 & \cellcolor{green!25}0.056 & \cellcolor{green!25}0.120 & \cellcolor{red!25}0.050 & \cellcolor{yellow!25}0.105 & \cellcolor{red!25}0.067 \\
\hline
\end{tabular}}
\end{threeparttable} 
\caption {{Tracking results on 7 challenging sequences in the Openloris-Scene \cite{shi2020openloris} dataset(RGB-D).  We also present the average tracking Completion Rate of 7 scenes. The best results are highlighted in red, the second-best results are highlighted in green, and the third-best results are highlighted in yellow. The evaluation was performed using the computation tools provided by the authors.}}\label{tab1}      
\end{table*}

\noindent\textbf{ScanNet \cite{dai2017scannet} Dataset Experimental Results. }Our evaluation results are based on the ScanNet\cite{dai2017scannet} dataset(RGB-D), which contains large-scale scenes. We assessed the average running speed, memory usage, mapping time, and GPU memory usage across six scenes. The results indicate that our method achieves higher running speed and reduces memory consumption. 
\begin{table}[h]
\centering 
\begin{center}
\begin{threeparttable}
\renewcommand\arraystretch{1.3} 

\useunder{\uline}{\ul}{}
\setlength{\tabcolsep}{3.5mm}
\begin{tabular}{c|ccccc}\hline
                                                      Method     &Mapping(ms)↓      &FPS↑  & \#param.↓  & GPU(G)↓  \\ \hline

iMAP { \cite{sucar2021imap}}                                      &360.8  &2.0   &\textbf{0.3}M  &8.7          \\ 
NICE-SLAM { \cite{zhu2022nice}}                                     &620.3    &0.6  &12.2M & 14.6      \\  
Co-SLAM { \cite{zhu2022nice}}                                     & {202.2}  &6.4   &\underline{0.8}M    &\underline{8.2}   \\
GO-SLAM { \cite{zhang2023goslam}}                                     & \underline{61.9}  &\underline{10.5}  &N/A    &15.6   \\
\hline
DDN-SLAM                                                  &\textbf{50.1}   & \textbf{20.4}   &19.3M    &\textbf{7.8}    \\  \hline

\end{tabular}
\end{threeparttable}   
\end {center}    
\caption{\textbf{Runtime analysis on the ScanNet \cite{dai2017scannet} dataset.} The best result is shown in bold, and the second-best result is indicated with an underline. The results of methods are reported based on the average of 6 scenes.}\label{tab2}
\end{table}

\section{Conclusion}
We propose DDN-SLAM, a neural implicit semantic SLAM system that enables stable tracking in complex occlusion and challenging environments. Through 2D-3D scene understanding, it achieves dense indoor scene reconstruction. Our system is more aligned with real-world requirements, capable of flexibly supporting various inputs while demonstrating superior performance. It addresses occlusion from complex moving objects, retains and reconstructs potential moving objects in the scene, all while balancing computational memory and running speed. Our experiments demonstrate that our method effectively avoids interference from severe occlusion on mapping and tracking, and achieves state-of-the-art performance on various datasets.

%
%
\end{document}